\documentclass{svproc}
\usepackage{fancyhdr}
\usepackage{url}
\usepackage{amssymb}
\usepackage{amsmath}
\usepackage{graphicx}

\usepackage{hyperref}
\usepackage[numbers,sort]{natbib}

\fancypagestyle{firstpage}{
    \fancyhead{} 
    \fancyfoot{} 
    \lfoot{$\dagger$ Corresponding authors} 
}

\begin{document}

\mainmatter              
\title{Decomposed Human Motion Prior for Video Pose Estimation via Adversarial Training}
\titlerunning{Wenshuo Chen}  
%
\author{Wenshuo Chen\inst{1} \and Xiang Zhou\inst{1} \and
Zhengdi Yu\inst{2} \and Weixi Gu \inst{3\dagger} \and Kai Zhang\inst{1,4\dagger}}
\authorrunning{Wenshuo Chen et al.} 
%
%
\institute{Tsinghua Shenzhen International Graduate School, Tsinghua Unversity, China\\
\email{\{cws21, zhoux21\}@mails.tsinghua.edu.cn} \\
\email{zhangkai@sz.tsinghua.edu.cn}
\and
Durham University, UK \\
\email{zhengdi.yu@durham.ac.uk} 
\and 
China Academy of Industrial Internet, China \\
\email{guweixigavin@gmail.com}
\and
Research Institute of Tsinghua, Pearl River Delta
}

\maketitle              

\begin{abstract}
Estimating human pose from video is a task that receives considerable attention due to its applicability in numerous 3D fields. The complexity of prior knowledge of human body movements poses a challenge to neural network models in the task of regressing keypoints. In this paper, we address this problem by incorporating motion prior in an adversarial way. Different from previous methods, we propose to decompose holistic motion prior to joint motion prior, making it easier for neural networks to learn from prior knowledge thereby boosting the performance on the task. We also utilize a novel regularization loss to balance accuracy and smoothness introduced by motion prior. Our method achieves 9\% lower PA-MPJPE and 29\% lower acceleration error than previous methods tested on 3DPW. The estimator proves its robustness by achieving impressive performance on in-the-wild dataset.
\end{abstract}
\section{Introduction}
\label{sec:intro}
It is essential to estimate accurate human pose from video due to its wide applications in AR/VR and motion capture industry. For years, researchers have tried to tackle the challenge by applying the powerful CNN architecture \cite{kocabas2020vibe, rempe2021humor,yuan2022glamr}. 

While existing methods \cite{kanazawa2018end,kolotouros2019learning} have demonstrated impressive ability in retrieving precise human pose from images, they fail to produce smooth and accurate pose from videos. Several methods like \cite{kocabas2020vibe,luo20203d,rempe2021humor,yuan2022glamr} tried to address such problems by subtly incorporating human motion prior learnt from mass motion datasets \cite{li2021ai, mahmood2019amass}. 
Regrettably, when estimating videos containing specific actions such as dancing, these methods may not yield the expected results. Driven by the significance, a question arises:\textit{ What is actually being learned by neural networks when learning motion prior?}

\thispagestyle{firstpage}

\newpage

When a neural network is learning prior of the human motion, such as walking, it needs to not only model the physical laws such as Newton's laws, but also model human habits. This habit is usually reflected in the way humans coordinate their bodies such as the fact that humans do not walk with the same hand and foot. However, in video human pose estimation task, modeling the relationship of human body is highly unnecessary, because most human pose estimation networks have predicted the approximate pose in a single frame. Based on the above reasons, we argue that holistic motion prior should be decomposed.

\begin{figure}
    \centering
    \centerline{\includegraphics[width=0.7\columnwidth]{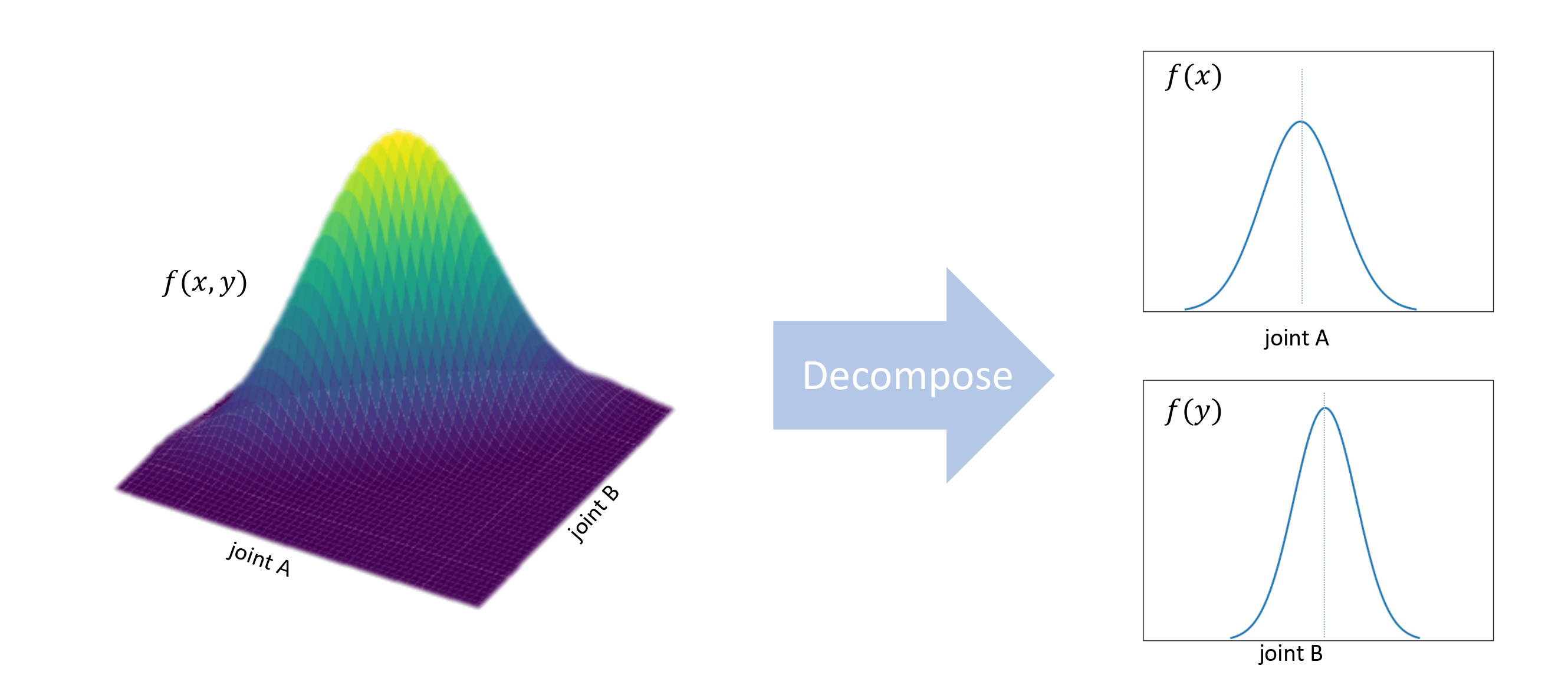}}
    \caption{The motion distribution of the holistic body can be viewed as a joint distribution of the motion distributions of each joint. Decomposing the relationship of joints will decrease the complexity of motion prior distribution.}
\end{figure}

In order to design a neural network for decomposing human body movements, we have designed independent temporal encoders for different joints. The joint-based approach that can significantly reduce the complexity of the holistic motion prior distribution, thus making it easier for neural networks to learn. Taking inspiration from VIBE \cite{kocabas2020vibe}, we also uses adversarial training to incorporate motion prior. As for avoiding the estimator to produce over smooth results, we propose a novel regularization loss to maintain the accuracy of the model.

As with most 3D human pose estimation methods, the output of our video pose estimator (generative model) is a pose sequence $\theta$ and shape sequence $\beta$ which determine a human motion sequence represented by SMPL model \cite{loper2015smpl}. The results and the real motion sequences sampled from the AMASS dataset \cite{mahmood2019amass} are continuously fed into the discriminator. With adversarial training and regularization loss, the joint-based recurrent neural network is able to learn motion prior and temporal information under the accuracy constraint.

The key contributions of this paper are summarized as follows: First, we propose a joint-based adversarial training method which decomposes complex human motion prior and produces smooth pose sequence from RGB video. Second, we propose a novel regularization loss to balance smoothness and accuracy. Third, we achieve much more reliable results than baseline and near state-of-the-art results on major benchmarks. Fourth, our key idea, decomposed motion prior, may inspire other tasks like motion synthesis and action recognition.

\section{Method}
\subsection{Preliminaries}
In order to describe the thousands of vertices around the human body under a certain pose, the most commonly used in academia and industry is the SMPL model which is also used in this paper. A SMPL model is controlled by two parameters, namely a pose paramter $\theta \in \mathbb{R}^{24 \times 3}$ which consists of rotation vectors of the 24 joints in an axis-angle representation, and a shape parameter $\beta \in \mathbb{R}^{10}$. They determine pretrained 3D human body $M(\theta, \beta)$. 

We aim to regress a sequence of SMPL parameters consisting of $\theta$ and $\beta$ from the input RGB video. 

\subsection{Modeling Motion Prior}
We are interested in modeling human motion prior because they are conducive for retrieving robust pose estimation from video. Previous state-of-the-art methods usually model motion distribution $p_{\phi}(\theta)$ as a whole, where $\theta$ is composed of a sequence of 24 joint poses $\left[\theta_{0}, \theta_{1},... \theta_{23} \right]$ and $\phi$ is a learnable parameter. Here we argue that holistic motion prior should be decomposed as motion prior for each joint, the decomposition reduces the complexity for the modeling of motion distribution thereby improves the overall performance. Because we focus on the distribution of motion for each joint, we model them by $p_{\phi_i}(\theta_i)$ and design independent temporal encoder for each joint where each encoder has its own parameters $\phi_i$.

\subsection{Combining Decomposed Motion Prior}
We choose PARE based on HRNet \cite{wang2020deep} as our backbone, PARE is trained with joint level supervision so that it can extract feature $f_i$ of joint $i$ and camera shape feature $c_i$ from input. We are able to implement a joint-based adversarial training to decompose motion prior with PARE conveniently.

A major challenge here is that we lack RGB videos with 3D supervsion and motion datasets with RGB information, the lacking of both forbids the use of supervised learning. Therefore we use adversarial learning to tackle the challenge. In adversarial training, the discriminator continually distinguishes samples from motion dataset from the output of RGB video pose estimator (generator), the estimator network gradually learns to predict results close to the motion groundtruths during training.

An overview of the model architecture is shown in Fig. 2. In the generator, the partial backbone outputs two feature volumes: one for the temporal branch to combine motion information and the other for the camera shape branch to predict weak camera and human shape. The temporal branch consists of temporal encoders and regressor heads. We expect each temporal encoder to learn the corresponding joint motion prior, so we separate it into 24 2-layer unidirectional GRUs. The features combined with temporal information is calculated as:
\begin{equation}
    \tilde{f}_i=GRU(f_i) .
\end{equation}

\begin{figure*}[t]
    \centering
    \includegraphics[width=1 \linewidth]{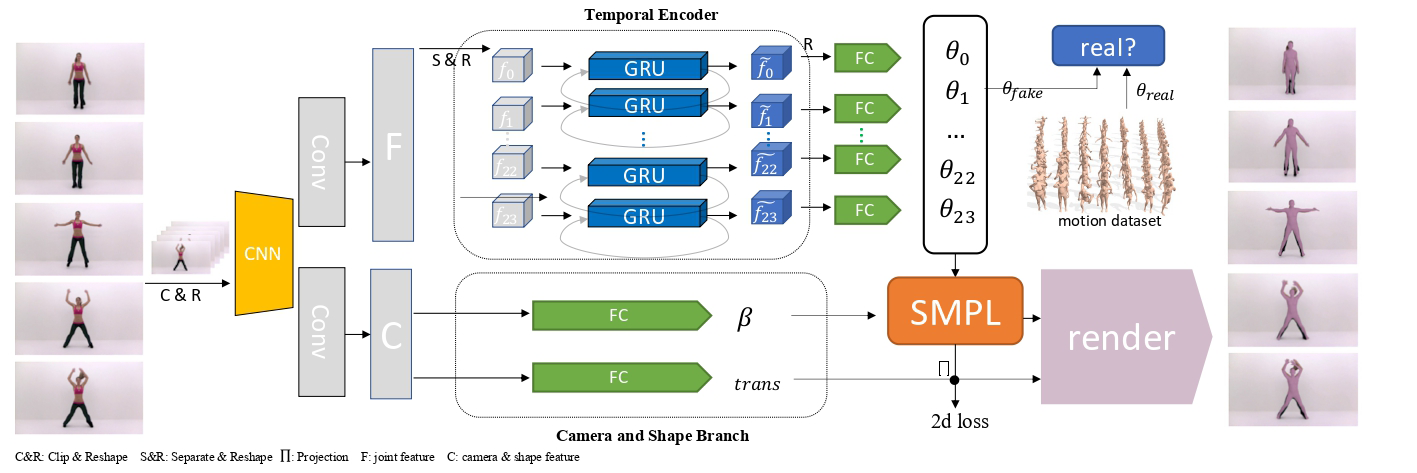}
    \caption{The figure illustrates that the partial backbone initially extracts features for each joint, followed by the use of independent GRUs designed to learn the motion prior of each joint, thereby forming the holistic motion prior. The discriminator continually distinguishes between samples from the motion dataset and the generator's output, ensuring that the estimator predicts more reliable results by combining the learned motion prior.}
    \label{fig:figure2}
\end{figure*}

The regressor head consists of 24 independent fully-connected layers $W_i$ for joint $i$. The output of each joint regressor head is then used to predict the joint pose: $\tilde{\theta}_i=W_i \tilde{f_i}$, where $\tilde{\theta_i}$ is the 6-D joint rotation representation we used to help train our networks. We use the 6-D representation because the matrix or vector representation is  noncontinuous in real Euclidean space while the 6-D representation is continuous \cite{zhou2019continuity}. Following \cite{zhou2019continuity} and Rodrigues's Formula, we can get the final pose estimation
\begin{equation}
\label{eq: diffusion process}
\begin{split}
& \theta_i=arccos(\frac{tr([\breve{\theta}_{i,0}, \breve{\theta}_{i,1}, \breve{\theta}_{i,2}]) - 1}{2}) , \\
\end{split}
\end{equation}
where
\begin{equation}
\label{6d 2 3x3}
\begin{split}
& \breve{\theta}_{i,0} = N(\tilde{\theta}_{i,0}) \\
& \breve{\theta}_{i,1} = N(\tilde{\theta}_{i,1}-(\breve{\theta}_{i,0} \cdot \tilde{\theta}_{i,1}) \breve{\theta}_{i,0}) \\
& \breve{\theta}_{i,2} = \breve{\theta}_{i,0} \times \breve{\theta}_{i,1}
\end{split}
\end{equation}
and $tr(.)$ is the trace of matrix and $N(.)$ denotes normalization.


To recover the human movement in real world coordinate system (root translation), the camera shape branch predicts weak camera $c_w=[s,t_x,t_y]$ which is used to calculate root translation while the shape parameters are predicted by independent linear regressor: $\beta=W_\beta c$, where $c=Concat([c_0,c_1,...c_{23}])$ and $c_w =W_c c$, where  $s$ denotes scale relative to original image, $t_x, t_y$ denote root translation in x and y dimension and $W_\beta$, $W_c$ are learnable parameters. Based on weak camera, we can simulate the translation in z-dimension by $t_z=\frac{2 \cdot f}{Res \cdot s}$, where $Res$ denotes the resolution of input image and $f$ denotes preset focal length. Finally, our video pose estimator outputs a sequence pose parameters $\theta$, root translation $trans=[t_x,t_y,t_z]$ and shape parameters $\beta$.

Our discriminant networks consist of several simple linear architecture with attention mechanism same as VIBE \cite{kocabas2020vibe}.

\subsection{Isometric Constraint}

Our human pose estimation network predicts SMPL parameters, and the 3D keypoints can be computed from SMPL parameters by SMPL pretrained model $\hat{y}_{3D}=M(\theta,\beta)$. The 2D prediction can then be obtained by projecting 3D keypoints to 2D image plane:
\begin{equation}
    \hat{y}^{2D} = K(R \hat{y}^{3D} + trans) ,
\end{equation}
where $K$ denotes presumptive camera instrinsics and $R$ is an identity matrix.

Based on the above supervision, we can summarize the loss function of the generator $L_G$ as follows:
\begin{equation}
   L_G=\lambda_{3D} L_{3D} + \lambda_{2D} L_{2D} + \lambda_{smpl} L_{smpl} + \lambda_{adv} L_{adv} + \lambda_{reg} L_{reg} ,
\end{equation}
where $L_{3D}$, $L_{2D}$ and $L_{smpl}$ are MSE loss and $\lambda$ denotes corresponding weight. The adversarial loss is designed in accordance to W-GAN\cite{arjovsky2017wasserstein}.
\begin{equation}
    L_{\text{adv}} = \mathbb{E}_{\theta \sim p_{G}}[(D_M(\theta^{\text{fake}})-1)^2] .
\end{equation}

A major difficulty in combining estimation results with motion prior is to avoid the overfitting of the temporal encoder to motion prior, which makes it sacrifice accuracy of the prediction for smoothness. To achieve a balanced trade-off between accuracy and smoothness, we propose a novel regularization loss $L_{reg}$:
\begin{equation}
    L_{reg}={\lVert \tilde{f_i}-f_i\rVert}_F.
\end{equation}
This simple trick greatly improves the performance.

For discriminator $D(.)$, we optimize the discriminator at fixed intervals with the following constraint:
\begin{equation}
    L_{D_M}=\mathbb{E}_{\theta \sim p_R}[(D_M(\theta^{fake})-1)^2]+\mathbb{E}_{\theta \sim P_G}[D_M(\theta^{real})^2] ,
\end{equation}
where $\theta^{fake}$ is the predicted pose sequence and $\theta^{real}$ is sampled randomly from the motion dataset.

\subsection{Implementation Details}

For an input RGB video, we first clip it to RGB images by a sliding window of size $T=16$, then use an object detector \cite{redmon2018yolov3} to determine the bounding box of human. After cropping and patching to $224 \times 224$ resolution, we extract their features via a pretrained PARE. Finally, we get SMPL prediction and weak camera.

Our proposed temporal encoder consists of 24 independent decomposed motion encoders. Each decomposed motion encoder consists of 2 layer GRU designed for independent joint and its input dimension is 128 corresponding to PARE feature volume and the hidden layer size is 64. Then a linear layer reshapes the feature size to $128$. Finally, independent regressor heads predict pose $\theta$. For camera and shape branch, we use two linear layers to predict weak camera $c_w$ and shape $\beta$.



\begin{table}
\centering
\caption{Comparison to the state-of-the-arts}
\label{table}
\small
\setlength{\tabcolsep}{3pt}
\begin{tabular}{|c|c|c|c|c|c|}
\hline
Method & 
PA-MPJPE & 
MPJPE & ACC & PVE & ACC ERR \\
\hline
HMR-temporal \cite{kanazawa2018end} & 
76.7 & 
130.0 & 
37.4 & - & - \\
\hline
SPIN-temporal \cite{kolotouros2019learning} & 
61.1 & 
102.4 & 
29.2 & 129.2 & 30.0 \\
\hline
VIBE \cite{kocabas2020vibe} & 
56.5 & 
93.5 & 
27.1 & 113.4 & 24.3 \\
\hline
\textbf{Ours} & 
\textbf{51.4} & 
\textbf{89.4} & 
\textbf{15.6} & 
\textbf{112.6} & 
\textbf{17.1} \\
\hline
\end{tabular}
\label{tab1}
\end{table}

\section{Experiments}
\subsection{Training Details}

MPII3D\cite{mehta2017monocular} and 3DPW \cite{von2018recovering} with 3D-2D groundtruth and InstaVariety\cite{kanazawa2019learning} with pseudo-2D groundtruth labelled by OpenPose\cite{cao2017realtime} are used for training. Adam\cite{kingma2014adam} with learning rate 0.0001 and weight decay 0.0001 is used to optimize our networks. To balance each isometric constraint, we set $\lambda_{3D}=300, \lambda_{2D}=300, \lambda_{SMPL}^{pose}=60, \lambda_{SMPL}^{\beta}=0.06$ and $L_{reg}=60$. We update the weights of the discriminator every 5 iterations and update generator's weights each iteration.

As we are most concerned with the accuracy and smoothness of the prediction, we set quantitative metrics for evaluating the two targets. The metric we use for precision is MPJPE (Mean Per Joint Position Error) calculated by mean position error for all joints. PA-MPJPE (Procrustes Alignment Mean Per Joint Position Error) is MPJPE under predicted keypoints aligned in 3D space. Acceleration (ACC) represents second order rate change of 3D keypoints in adjacent frames and Acceleration error (ACC ERR) is the error of ACC. They are two measurements of smoothness.

Our method is mainly based on the baseline VIBE, hence the comparison with VIBE can directly show the improvement from our method. VIBE is also based on the adversarial training to boost the temporal encoder to combine with motion prior. However, our method works better on accuracy and smoothness. We train our decomposed temporal encoder on MPII3D and InstaVariety, and test on the 3DPW test set, our method gets 9.0\% and 29.6\% improvement on accuracy and smoothness (See Table I). The performance far surpasses that of single-frame video pose estimation, which proves our method is robust to video pose estimation. Fig. 3. demonstrates our method is able to predict smooth and high quality human pose. And Fig. 4. proves our method can achieve satisfying results even in scenes with occlusion.

\subsection{Evaluation}

As we are most concerned with the accuracy and smoothness of the prediction, we set quantitative metrics for evaluating the two targets. The metric we use for precision is MPJPE (Mean Per Joint Position Error) calculated by mean position error for all joints. PA-MPJPE (Procrustes Alignment Mean Per Joint Position Error) is MPJPE under predicted keypoints aligned in 3D space. Acceleration (ACC) represents second order rate change of 3D keypoints in adjacent frames and Acceleration error (ACC ERR) is the error of ACC. They are two measurements of smoothness. 

\begin{figure}
    \centering
    \centerline{\includegraphics[width=0.8\columnwidth]{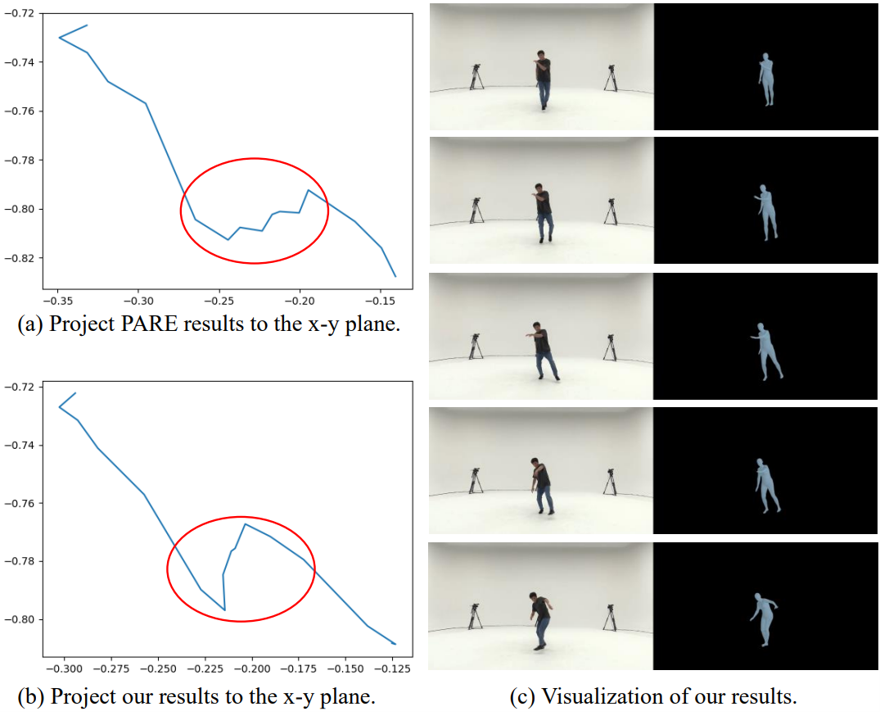}}
    \caption{Visualize our results on the hm3.6 dataset \cite{ionescu2013human3}. Figure (c) is the visualization of our results extracted at every 8 frames on the original image right, which shows plentiful motion. Figures (a) and (b) are the first 16 frame right shoulder coordinates predicted by PARE and ours on the x-y plane, respectively. Figure (a) shows a trigonometric-like jitter.}
\end{figure}

Our method is mainly based on the baseline VIBE, hence the comparison with VIBE can directly show the improvement from our method. VIBE is also based on the adversarial training to boost the temporal encoder to combine with motion prior. However, our method works better on accuracy and smoothness. We train our decomposed temporal encoder on MPII3D and InstaVariey, and test on the 3DPW test set, our method gets 9.0\% and 29.6\% improvement on accuracy and smoothness (See Table I). The performance far surpasses that of single-frame video pose estimation, which proves our method is robust to video pose estimation. Fig. 3. demonstrates our method is able to predict smooth and high quality human pose. And Fig. 4. proves our method can achieve satisfying results even in scenes with occlusion.
\begin{figure}
    \centering
    \centerline{\includegraphics[width=0.7\columnwidth]{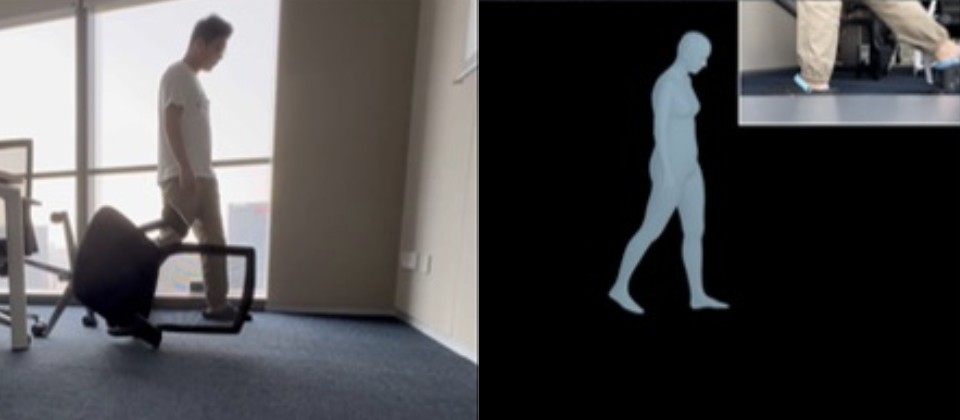}}
    \caption{Even in the presence of occlusion, our method can still infer the correct human pose using temporal information and motion prior.}
\end{figure}

\begin{table} 
\centering
\caption{Ablation Study. sep.t mmeans that we use independent temporal encoder to decompose human motion.  Decomposing the joints is beneficial for better incorporating motion priors, which reduces jitters but lead to the model being biased towards producing smoother rather than accurate results. The regularization loss function we designed avoids this phenomenon.}
\label{table}
\small
\setlength{\tabcolsep}{3pt}
\begin{tabular}{|c|c|c|c|c|}
\hline
Method& 
PA-MPJPE& 
MPJPE&
ACC&
ACC ERR\\
\hline
PARE& 
69.9&
106.3&
30.0&
29.9\\
\hline
PARE+sep.t& 
86.1&
122.1&
22.8&
23.4\\
\hline
\textbf{PARE+sep.t+$L_{reg}$}& 
\textbf{68.39}&
\textbf{104.2}&
\textbf{21.1}&
\textbf{21.5}\\
\hline
\end{tabular}
\label{tab1}
\end{table}


\subsection{Ablation Study}

Table II shows the results from our ablation experiments on the MPII3D test set. As the second row of the table shows, temporal encoder like GRU can do harm to accuracy because it may overfit to AMASS dataset and destroys spatial information. In spite of a drop in ACC and ACC ERROR, the accuracy of the predictions also decreases. The third row of the table shows the proposed novel regularization loss function can keep a balance between motion priors and accuracy. Our method benefit from decomposed motion prior and regularization loss to achieve high performance on video human pose estimation.

\section{Conclusion}

In this paper, we propose a robust video human pose estimation approach combining human motion prior via adversarial training. We model human motion as distribution of each joint motion and design joint-based temporal encoder to decompose the complex motion prior. We also propose a novel regularization loss to avoid the destruction of spatial information to maintain the accuracy of estimator. Decomposed motion prior greatly benefits video human pose estimation and we believe it will give great inspiration to motion synthesis and action recognition tasks.

\section{Acknowledgement}
This work was supported by the key-Area Research and Development Program of Guangdong Province (2020B0909050003).

\end{document}